\documentclass[sigconf]{acmart}
\usepackage{subfigure}
\usepackage{multirow}
\usepackage{pifont}
\usepackage{bbding}
\usepackage{adjustbox}
\usepackage{threeparttable}
\usepackage{colortbl}
\usepackage{xcolor}

\usepackage{bbm}

\usepackage{framed}
\definecolor{shadecolor}{RGB}{240,240,240}

\newcommand{\vpara}[1]{\vspace{0.05in}\noindent\textbf{#1 }}

\AtBeginDocument{%
  }

\copyrightyear{2026}
\acmYear{2026}
\setcopyright{cc}
\setcctype{by-nc-nd}
\acmConference[WWW '26]{Proceedings of the ACM Web Conference 2026}{April 13--17, 2026}{Dubai, United Arab Emirates}
\acmBooktitle{Proceedings of the ACM Web Conference 2026 (WWW '26), April 13--17, 2026, Dubai, United Arab Emirates}
\acmPrice{}
\acmDOI{10.1145/3774904.3792965}
\acmISBN{979-8-4007-2307-0/2026/04}

\acmSubmissionID{123-A56-BU3}

\begin{document}

\title{Tabular Foundation Models are Strong Graph Anomaly Detectors}

\settopmatter{authorsperrow=4}

\author{Yunhui Liu}
\orcid{0009-0006-3337-0886}
\affiliation{
\institution{State Key Laboratory for Novel Software Technology\\ Nanjing University}
\city{Nanjing}
\country{China}}
\affiliation{
\institution{Ant Group}
\city{Hangzhou}
\country{China}}
\email{lyhcloudy1225@gmail.com}

\author{Tieke He}
\authornote{Corresponding authors.}
\orcid{0000-0001-9649-1796}
\affiliation{
\institution{State Key Laboratory for Novel Software Technology\\ Nanjing University}
\city{Nanjing}
\country{China}}
\email{hetieke@gmail.com}

\author{Yongchao Liu}
\authornotemark[1]
\orcid{0000-0003-3440-9675}
\author{Can Yi}
\orcid{0009-0002-8886-7276}
\affiliation{%
\institution{Ant Group}
\city{Hangzhou}
\country{China}}
\email{yongchao.ly@antgroup.com}
\email{yican.yc@antgroup.com}

\author{Hong Jin}
\orcid{0009-0000-8277-6387}
\author{Chuntao Hong}
\orcid{0009-0009-3472-6102}
\affiliation{%
\institution{Ant Group}
\city{Hangzhou}
\country{China}}
\email{jinhong.jh@antgroup.com}
\email{chuntao.hct@antgroup.com}

\renewcommand{\shortauthors}{Yunhui Liu et al.}

\begin{abstract}
Graph anomaly detection (GAD), which aims to identify abnormal nodes that deviate from the majority, has become increasingly important in high-stakes Web domains.
However, existing GAD methods follow a "one model per dataset" paradigm, leading to high computational costs, substantial data demands, and poor generalization when transferred to new datasets.
This calls for a foundation model that enables a "one-for-all" GAD solution capable of detecting anomalies across diverse graphs without retraining.
Yet, achieving this is challenging due to the large structural and feature heterogeneity across domains.
In this paper, we propose TFM4GAD, a simple yet effective framework that adapts tabular foundation models (TFMs) for graph anomaly detection.
Our key insight is that the core challenges of foundation GAD, handling heterogeneous features, generalizing across domains, and operating with scarce labels, are the exact problems that modern TFMs are designed to solve via synthetic pre-training and powerful in-context learning.
The primary challenge thus becomes structural: TFMs are agnostic to graph topology.
TFM4GAD bridges this gap by "flattening" the graph, constructing an augmented feature table that enriches raw node features with Laplacian embeddings, local and global structural characteristics, and anomaly-sensitive neighborhood aggregations.
This augmented table is processed by a TFM in a fully in-context regime.
Extensive experiments on multiple datasets with various TFM backbones reveal that TFM4GAD surprisingly achieves significant performance gains over specialized GAD models trained from scratch.
Our work offers a new perspective and a practical paradigm for leveraging TFMs as powerful, generalist graph anomaly detectors.
Our code can be found at \url{https://github.com/Cloudy1225/TFM4GAD}.
\end{abstract}

\ccsdesc[500]{Computing methodologies~Machine learning}

\keywords{Graph Anomaly Detection, Tabular Foundation Model, Graph Foundation Model}


\maketitle

\section{Introduction}
Graph anomaly detection (GAD), which aims to identify abnormal nodes that significantly deviate from the majority, has attracted growing attention in high-stakes Web domains such as finance, social networks, and cybersecurity~\cite{GADSurvey}. 
Despite recent progress, existing GAD methods typically follow a "one model per dataset" paradigm, requiring dataset-specific training and sufficient training data to build an effective detector for each graph. 
This paradigm suffers from three inherent limitations~\cite{ARC}: 
1) \textit{Expensive training cost.} Each dataset requires training a GAD model from scratch, which is computationally intensive, especially for large-scale graphs.
2) \textit{High data requirements.} Reliable model training demands sufficient in-domain data and labels, posing challenges in scenarios with sparse data, privacy restrictions, or expensive anomaly labeling.
3) \textit{Limited generalizability.} Adapting to a new dataset typically involves hyperparameter tuning or architectural adjustments, further increasing deployment cost and hindering scalability across domains.

Therefore, there is a pressing need for foundation GAD models that can be trained once and directly generalize across diverse graphs.
However, developing such models is inherently challenging.
Graphs originate from heterogeneous domains (e.g., social, financial, or co-purchasing networks) where both structural patterns and node features (in dimensionality and semantics) vary drastically. 
Recent attempts~\cite{ARC, UNPrompt, AnomalyGFM} employ simple dimensionality reduction methods such as PCA to project node features into a fixed-size space. 
Yet, such transformations fail to effectively exploit arbitrary node features from new domains and still rely heavily on labeled supervision.
In practice, anomaly labels are scarce, costly to obtain, and fundamentally limited~\cite{ConsisGAD, SpaceGNN, APF}.

In this work, we propose a different approach, motivated by a key insight: the node features in many GAD scenarios are \textit{tabular} in nature.
They often consist of numerical or categorical attributes, such as transaction statistics or temporal descriptors.
Even when node features are textual, standard methods like TF-IDF or sentence embeddings naturally convert them into a numerical tabular format.
This suggests that the challenge of feature heterogeneity in GAD is directly analogous to the primary challenge addressed by tabular foundation models (TFMs)~\cite{TabPFN, TabPFNv2}.
Based on this, instead of building a complex, graph-native foundation model, we leverage the remarkable capabilities of TFMs.
We argue that TFMs are exceptionally well-suited to the foundation GAD problem for three reasons.
First, they are true \textit{in-context learners}, designed to approximate Bayesian inference and make predictions on new tasks without retraining, directly addressing the limitations of high cost and poor generalizability.
Second, TFMs treat features as a set of tokens, making them inherently robust to the heterogeneous feature dimensionalities and semantics that plague cross-domain GAD.
Third, TFMs are pre-trained on vast synthetic priors, mitigating the critical reliance on scarce, real-world anomaly labels.

The primary challenge, therefore, becomes a structural one: TFMs are agnostic to the graph topology.
We bridge this gap with TFM4GAD, a framework that "flattens" graph data into a tabular representation that is both structure-aware and anomaly-sensitive.
TFM4GAD constructs an augmented feature table where each node's feature vector is enriched with four key components:
(1) its raw node features;
(2) global positional encodings via Laplacian embeddings;
(3) local and global structural characteristics, such as Node Degree and PageRank scores; and
(4) an anomaly-sensitive neighborhood aggregation based on Beta Wavelet filters~\cite{BWGNN}, which are specifically designed to capture high-frequency deviations and avoid the over-smoothing that obscures anomalies.
This augmented table is then processed directly by a TFM. The TFM uses a small set of known labels as in-context examples to predict anomaly scores for all unlabeled nodes.
We conduct experiments on four benchmark datasets with four TFM backbones and find that this surprisingly simple framework significantly outperforms strong GAD-specific models trained from scratch.
These results highlight that building foundation graph anomaly detectors on top of tabular foundation models is a highly promising research direction.

\section{Preliminary}
\subsection{Graph Anomaly Detection}
We define an attributed graph as $\mathcal{G} = (\mathcal{V}, \mathcal{E}, \boldsymbol{X})$, where $\mathcal{V} = \{v_1, \dots, v_n\}$ is the set of $n$ nodes, $\mathcal{E}$ is the edge set, and $\boldsymbol{X} \in \mathbb{R}^{n \times d}$ is the node feature matrix.
The goal of GAD is to assign an anomaly score or binary label $\boldsymbol{y}_v \in \{0, 1\}$ to each node $v$, where $\boldsymbol{y}_v = 1$ indicates an anomalous node and $\boldsymbol{y}_v = 0$ indicates a normal one.
A key challenge, which we focus on, is developing a \textit{foundation model} for GAD.
Such a model $F$ must generalize across diverse graph datasets (e.g., from finance, social networks, etc.) that vary significantly in structure $\mathcal{E}$ and feature semantics $\boldsymbol{X}$.
Given a small set of labeled nodes $\mathcal{D}_L = \{(\boldsymbol{x}_v, \boldsymbol{y}_v) \mid v \in \mathcal{V}_L\}$ from a new graph $\mathcal{G}$ as in-context examples, the model should be able to predict the labels $\hat{\boldsymbol{y}}_U$ for the remaining unlabeled nodes $\mathcal{V}_U$ \textit{without any training or fine-tuning}.


\subsection{Tabular Foundation Models}
A tabular dataset is a table $\mathcal{T} = (\boldsymbol{X}, \boldsymbol{y})$, where $\boldsymbol{X} \in \mathbb{R}^{n \times d}$ consists of $n$ instances (rows) and $d$ heterogeneous features (columns), which can be numerical or categorical.
The label vector $\boldsymbol{y}$ provides supervision for a downstream task.
A tabular foundation model (TFM) is a pre-trained model $F_{\theta}$ designed to learn a universal inference mechanism applicable to new, unseen tabular datasets $\mathcal{T}_{\text{new}}$ without dataset-specific training.
Leading TFMs~\cite{TabPFN, TabPFNv2, LimiX, TabPFNv2.5} achieve this by learning to approximate the Bayesian posterior predictive distribution.
They are trained on a vast collection of synthetic datasets sampled from a prior, learning a "prior-data fitted network" (PFN)~\cite{PFNs}.
This enables them to perform in-context Bayesian inference:
$$
p(\hat{\boldsymbol{y}}_{\text{test}} \mid \boldsymbol{X}_{\text{test}}, \boldsymbol{X}_{\text{train}}, \boldsymbol{y}_{\text{train}}) \approx F_{\theta}(\boldsymbol{X}_{\text{test}}, \boldsymbol{X}_{\text{train}}, \boldsymbol{y}_{\text{train}})
$$
where $(\boldsymbol{X}_{\text{train}}, \boldsymbol{y}_{\text{train}})$ are provided as in-context examples at inference time.
Recent works~\cite{G2T-FM, TAG} have also shown the potential of TFMs in graph learning.


\section{Framework}
\begin{figure}[ht]
\centerline{\includegraphics[width=1.\linewidth]{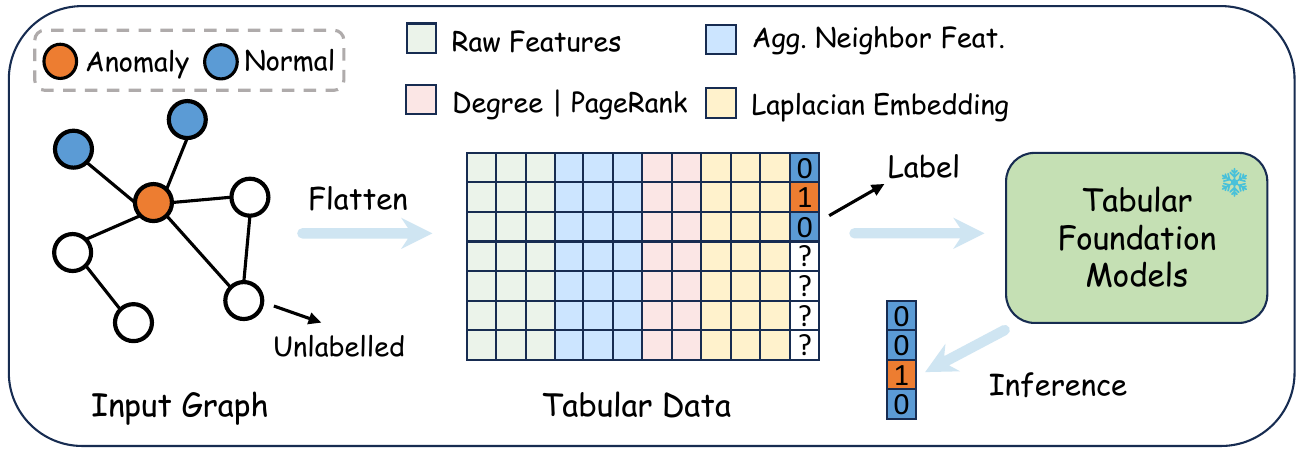}}
\caption{Overview of the TFM4GAD framework.} 
\label{fig:framework}
\vspace{-1em}
\end{figure}

In this section, we present TFM4GAD, as illustrated in Figure~\ref{fig:framework}.

\subsection{Graph-to-Tabular
Flattening}
The central challenge in applying TFMs to GAD is the structural gap: TFMs are agnostic to the graph's edge set $\mathcal{E}$.
A naive application using only the raw node features $\boldsymbol{X}$ would discard all topological information, which is critical for identifying structural anomalies.

To bridge this gap, TFM4GAD introduces a graph-to-table "flattening" process.
We construct an augmented feature matrix $\boldsymbol{X}_{\text{aug}} \in \mathbb{R}^{n \times d_{\text{aug}}}$ where each node $v \in \mathcal{V}$ is represented by a new, enriched feature vector $\boldsymbol{x}'_v$.
This vector is a concatenation of the original features and a set of engineered, anomaly-sensitive structural representations.
Specifically, the augmented feature vector $\boldsymbol{x}'_v$ for each node $v$ is defined as:
$$
\boldsymbol{x}'_v = [\boldsymbol{x}_v \mathbin\Vert \boldsymbol{x}_{\text{lap}} \mathbin\Vert \boldsymbol{x}_{\text{char}} \mathbin\Vert \boldsymbol{x}_{\text{nbr}}]
$$
where $\boldsymbol{x}_v$ is the raw feature, $\boldsymbol{x}_{\text{lap}}$ contains Laplacian embeddings, $\boldsymbol{x}_{\text{char}}$ captures explicit structural characteristics, and $\boldsymbol{x}_{\text{nbr}}$ provides a non-parametric neighborhood aggregation. 
Notably, TFMs typically apply feature shuffling and ensembling during inference to approximate permutation invariance~\cite{TabPFNv2, LimiX}, so the ordering of these concatenated feature blocks has little effect on performance.
We detail each component below.

\vpara{Raw Features and Laplacian Embeddings.}
The first component is the node's original feature vector $\boldsymbol{x}_v \in \mathbb{R}^{d}$.
To provide the TFM with a global understanding of node positions within the graph topology, we include Laplacian Embeddings as the second component, $\boldsymbol{x}_{\text{lap}}$.
These are computed as the $k$ eigenvectors corresponding to the smallest non-zero eigenvalues of the normalized graph Laplacian $\boldsymbol{L} = \boldsymbol{I} - \boldsymbol{D}^{-1/2} \boldsymbol{A} \boldsymbol{D}^{-1/2}$.
This component, $\boldsymbol{x}_{\text{lap}} \in \mathbb{R}^{k}$, serves as a form of positional encoding, mapping nodes to a low-dimensional space that preserves large-scale structural relationships.

\vpara{Local and Global Structural Characteristics.}
While Laplacian embeddings capture global topology, anomalies often manifest as sharp deviations in local or centrality-based patterns.
We therefore compute a set of explicit structural characteristics, $\boldsymbol{x}_{\text{char}}$.
This component includes fundamental node-level metrics known to be highly sensitive to anomalous behavior:
\begin{itemize}
    \item \textbf{Node Degree:} The degree $\text{deg}(v)$ of a node, which quantifies its local connectivity. Anomalies may present as extreme high-degree (e.g., spam bots) or low-degree (e.g., isolated fraudulent accounts) nodes.
    \item \textbf{PageRank Score:} The PageRank score $\text{PR}(v)$, which measures a node's global importance or influence. This metric can identify nodes that are strategically connected to other important nodes, a pattern often manipulated by sophisticated attackers.
\end{itemize}
These features provide the TFM with explicit, interpretable signals about a node's role and connectivity.

\vpara{Anomaly-Sensitive Neighborhood Aggregation.}
The components above describe a node in isolation or its global position. However, anomalous behavior is often defined by a node's immediate neighborhood.
Standard GNNs (e.g., GCN) act as low-pass filters, which suffer from over-smoothing.
This averaging effect is particularly detrimental to GAD, as it can "drown out" the unique signal of an anomalous node by blending it with its normal neighbors.

To counter this, we alternatively adopt a non-parametric neighborhood aggregation strategy based on Beta Wavelets, inspired by~\cite{BWGNN}.
This approach is designed to preserve, rather than smooth, anomaly-sensitive signals.
The core insight from~\cite{BWGNN} is that anomalies often manifest as high-frequency "right-shift" phenomena in the graph spectrum.
Beta wavelets, unlike low-pass filters, function as a bank of localized band-pass filters, enabling them to capture these high-frequency deviations.
A Beta wavelet transform $\mathcal{W}_{p,q}$ is defined as a polynomial of the normalized Laplacian $\boldsymbol{L}$:
$$
\mathcal{W}_{p,q} = \beta_{p,q}^{*}(\boldsymbol{L}) = \frac{1}{2B(p+1,q+1)} \left(\frac{\boldsymbol{L}}{2}\right)^{p} \left(\boldsymbol{I} - \frac{\boldsymbol{L}}{2}\right)^{q}
$$
where $p, q \in \mathbb{N}^{+}$ control the filter's band-pass characteristics and $B(\cdot)$ is the Beta function.
A key property of these filters is their ability to produce both positive and negative responses, allowing them to model both similarity and \textit{dissimilarity} within a neighborhood, thereby making anomalous nodes more distinguishable~\cite{BWGNN}.

We compute a bank of these wavelet-filtered features by applying $C+1$ filters (where $p+q=C$) to the raw node features $\boldsymbol{X}$:
$$
\boldsymbol{X}_{\text{nbr}} = \left[ \mathcal{W}_{0,C}(\boldsymbol{X}) \mathbin\Vert \mathcal{W}_{1,C-1}(\boldsymbol{X}) \mathbin\Vert \dots \mathbin\Vert \mathcal{W}_{C,0}(\boldsymbol{X}) \right]
$$
This provides a rich, multi-scale, non-parametric representation of each node's neighborhood. The resulting feature vector for a single node, $\boldsymbol{x}_{\text{nbr}}$ (i.e., the $v$-th row of $\boldsymbol{X}_{\text{nbr}}$), has a dimensionality of $d \cdot (C+1)$, and is highly sensitive to local deviations.

\subsection{In-Context Graph Anomaly Detection}
After constructing the final augmented feature matrix $\boldsymbol{X}_{\text{aug}}$ by concatenating all components for every node, TFM4GAD performs graph anomaly detection in a single, in-context inference pass.

Let $\mathcal{D}_{\text{train}} = (\boldsymbol{X}_{\text{aug}}[\mathcal{V}_L], \boldsymbol{y}_L)$ be the "in-context training set" derived from the set of labeled nodes $\mathcal{V}_L$, and let $\mathcal{D}_{\text{test}} = (\boldsymbol{X}_{\text{aug}}[\mathcal{V}_U])$ be the "test set" corresponding to the unlabeled nodes $\mathcal{V}_U$.
The TFM is then prompted to predict the anomaly probabilities for the unlabeled nodes:
$$
p(\hat{\boldsymbol{y}}_U \mid \mathcal{D}_{\text{train}}, \mathcal{D}_{\text{test}}) = \text{TFM}(\boldsymbol{X}_{\text{aug}}[\mathcal{V}_U], \boldsymbol{X}_{\text{aug}}[\mathcal{V}_L], \boldsymbol{y}_L)
$$
The TFM, having been pre-trained on a vast array of synthetic tabular tasks, approximates the Bayesian posterior predictive distribution.
It leverages the labeled examples $(\boldsymbol{X}_{\text{aug}}[\mathcal{V}_L], \boldsymbol{y}_L)$ as in-context "evidence" to infer the relationship between the augmented features and the anomaly labels.
It then applies this inferred model to the unlabeled nodes $\boldsymbol{X}_{\text{aug}}[\mathcal{V}_U]$ to predict their anomaly scores.
The entire TFM4GAD framework operates without any graph-specific training, fine-tuning, or gradient updates, enabling in-context generalization to new GAD datasets.

\section{Experiments}
\subsection{Experimental Setup}
\vpara{Datasets and Baselines.} 
We evaluate TFM4GAD on four representative datasets from GADBench~\cite{GADBench}: Amazon, YelpChi, T-Finance, and T-Social. 
These datasets cover diverse domains and differ in both feature dimensionality and graph scale. 
Detailed dataset statistics are summarized in Table~\ref{tab:dataset}.
We compare against a broad range of supervised baseline methods that require dataset-specific training: GCN~\cite{GCN}, AMNet~\cite{AMNet}, BWGNN~\cite{BWGNN}, GHRN~\cite{GHRN}, RFGraph~\cite{GADBench}, XGBGraph~\cite{GADBench}, ConsisGAD~\cite{ConsisGAD}, and SpaceGNN~\cite{SpaceGNN}.

\vpara{Metrics and Implementation Details.} 
Following GADBench~\cite{GADBench}, we report AUROC and AUPRC as evaluation metrics. 
To simulate realistic label scarcity, each dataset contains only 100 labeled nodes (including 20 anomalies). 
All results are averaged over 10 random splits provided by GADBench to ensure fair comparison.
For tabular foundation models, we adopt TabPFNv2~\cite{TabPFNv2}, TabPFNv2.5~\cite{TabPFNv2.5}, LimiX-2M, and LimiX-16M~\cite{LimiX} as backbones. 
We set $k=16$ eigenvectors for Laplacian embeddings $\boldsymbol{x}_{\text{lap}}$.
The hop order of neighborhood aggregation is selected from $\{1,2,3\}$.
Our code can be found at \url{https://github.com/Cloudy1225/TFM4GAD}.

\begin{table}[ht]
    \centering
    \setlength{\tabcolsep}{2pt}
    \caption{Dataset statistics.}
    \label{tab:dataset}
    \vspace{-1em}
    \begin{tabular}{l|ccccc}
        \toprule
        Dataset & Domain & \#Nodes & \#Edges & \#Dim. & Anomaly \\
        \midrule
        Amazon & Co-review & 11,944 & 4,398,392 & 25 & 9.5\% \\
        YelpChi & Co-review & 45,954 & 3,846,979 & 32 & 14.5\% \\
        T-Finance & Transaction & 39,357 & 21,222,543 & 10 & 4.6\% \\
        T-Social & Social & 5,781,065 & 73,105,508 & 10 & 3.0\% \\
        \bottomrule
    \end{tabular}
    \vspace{-1em}
\end{table}

\subsection{Experimental Results}

\begin{table*}[ht]
    \centering
    \setlength{\tabcolsep}{3pt}
    \caption{
    Performance comparison. Baseline results are from GADBench except ConsisGAD and SpaceGNN.
    }
    \vspace{-1em}
    \label{tab:performance}
    \begin{tabular}{l|l|cccccccc|cc}
        \toprule
        \multirow{2}{*}{} & \multirow{2}{*}{Method} 
        & \multicolumn{2}{c}{Amazon} & \multicolumn{2}{c}{YelpChi} 
        & \multicolumn{2}{c}{T-Finance} & \multicolumn{2}{c|}{T-Social} 
        & \multicolumn{2}{c}{Average} \\
        \cmidrule(lr){3-4} \cmidrule(lr){5-6} \cmidrule(lr){7-8} \cmidrule(lr){9-10} \cmidrule(lr){11-12}
        & & AUROC & AUPRC & AUROC & AUPRC & AUROC & AUPRC & AUROC & AUPRC & AUROC & AUPRC \\
        \midrule

        \multirow{8}{*}{\rotatebox{90}{Traditional}} 
        & GCN & 82.0±0.3 & 32.8±1.2 & 51.2±3.7 & 16.4±2.6 & 88.3±2.5 & 60.5±10.8 & 71.6±10.4 & 8.4±3.8 & 73.28 & 29.52 \\
        & AMNet & 92.8±2.1 & 82.4±2.2 & 64.8±5.2 & 23.9±3.5 & 92.6±0.9 & 60.2±8.2 & 53.7±3.4 & 3.1±0.3 & 75.97 & 42.40 \\
        & BWGNN & 91.8±2.3 & 81.7±2.2 & 64.3±3.4 & 23.7±2.9 & 92.1±2.7 & 60.9±13.8 & 77.5±4.3 & 15.9±6.2 & 81.42 & 45.55 \\
        & GHRN & 90.9±1.9 & 80.7±1.7 & 64.5±3.1 & 23.8±2.8 & 92.6±0.7 & 63.4±10.4 & 78.7±3.0 & 16.2±4.6 & 81.68 & 46.02 \\
        & RFGraph & 92.5±1.3 & 70.7±5.1 & 61.6±2.7 & 23.6±2.5 & 95.0±0.7 & 81.1±2.7 & 88.6±1.6 & 51.3±6.2 & 84.42 & 56.68 \\
        & XGBGraph & 94.7±0.9 & 84.4±1.1 & 64.0±3.5 & 24.8±3.1 & 94.8±0.6 & 78.3±3.1 & 85.2±1.8 & 40.6±7.6 & 84.68 & 57.02 \\
        & ConsisGAD & 92.31±2.16 & 78.68±5.65 & 66.08±3.80 & 25.85±2.93 & 94.31±0.76 & 79.68±4.73 & 93.07±1.90 & 41.29±4.97 & 86.44 & 56.38 \\
        & SpaceGNN & 91.08±2.50 & 81.14±2.29 & 66.81±2.75 & 25.73±2.35 & 93.41±0.96 & 81.03±3.46 & 94.72±0.65 & 58.99±5.69 & 86.50 & 61.72 \\
        \midrule

        \multirow{4}{*}{\rotatebox{90}{TFM4GAD}} 
        & LimiX-2M & 95.65±1.25 & 83.67±5.19 & 70.83±4.36 & \textbf{31.96±4.88} & \underline{95.69±0.40} & \underline{83.49±2.66} & 96.44±1.01 & 83.26±4.33 & 89.65 & 70.60 \\
        & LimiX-16M & \textbf{96.14±1.02} & 85.50±4.24 & 70.54±4.93 & 31.57±5.16 & \textbf{96.01±0.40} & \textbf{84.92±1.63} & \underline{96.98±0.63} & \underline{87.01±2.18} & \textbf{89.92} & \textbf{72.25} \\
        & TabPFNv2 & 95.94±1.03 & \underline{86.76±1.91} & \underline{70.91±4.92} & 30.72±4.89 & 95.38±0.66 & 82.75±2.75 & 96.93±0.60 & 86.14±3.08 & 89.79 & \underline{71.59} \\
        & TabPFNv2.5 & \underline{96.12±0.99} & \textbf{87.28±1.61} & \textbf{71.64±3.62} & \underline{31.69±4.13} & 94.52±1.04 & 79.55±3.37 & \textbf{97.16±0.56} & \textbf{87.19±2.48} & \underline{89.86} & 71.43 \\
        \bottomrule
    \end{tabular}
\end{table*}

\vpara{Performance Comparison.}
As shown in Table \ref{tab:performance}, TFM4GAD significantly outperforms all specialized GAD baselines on both average AUROC and AUPRC metrics. 
This result is particularly compelling as TFM4GAD operates in a purely in-context learning regime, requiring no graph-specific training or gradient updates. 
Our best-performing variant (LimiX-16M) achieves an average AUROC of 89.92\% and AUPRC of 72.25\%, decisively surpassing the strongest trained baseline, SpaceGNN (86.50\% AUROC and 61.72\% AUPRC). 
This performance gap is consistent across individual datasets, such as T-Social and YelpChi, demonstrating the efficacy of our structure-aware "flattening" approach in transforming TFMs into powerful, generalist graph anomaly detectors.

\vpara{Ablation Study.}
We analyze the contribution of each feature component in Table \ref{tab:ablation}, reporting AUPRC on T-Finance and T-Social.
Using only raw features ($\boldsymbol{x}_v$) performs poorly, showing that structure is essential. The largest performance gain comes from adding the anomaly-sensitive neighborhood aggregation ($\boldsymbol{x}_{\text{nbr}}$). Adding explicit structural characteristics ($\boldsymbol{x}_{\text{char}}$) provides another significant boost. Finally, Laplacian embeddings ($\boldsymbol{x}_{\text{lap}}$) offer a final refinement. This confirms that all components of our graph-to-tabular flattening process are crucial for the model's high performance.

\begin{table}[ht]
    \centering
    \setlength{\tabcolsep}{2pt}
    \caption{Ablation study.}
    \label{tab:ablation}
    \vspace{-1em}
    \resizebox{0.49\textwidth}{!}{
    \begin{tabular}{l|cccc}
        \toprule
        \multirow{2}{*}{Variant} & \multicolumn{2}{c}{T-Finance} & \multicolumn{2}{c}{T-Social}\\
        \cmidrule(lr){2-3} \cmidrule(lr){4-5}
                               & LimiX-2M & TabPFNv2 & LimiX-16M & TabPFNv2.5 \\
        \midrule
        raw features $\boldsymbol{x}_v$ & 76.23±0.87 & 75.76±1.50 & 7.41±1.61 & 7.52±1.67 \\
        + $\boldsymbol{x}_{\text{nbr}}$ & 82.87±2.48 & 82.31±2.10 & 77.99±4.40 & 78.11±7.20 \\
        + $[\boldsymbol{x}_{\text{char}} \mathbin\Vert \boldsymbol{x}_{\text{nbr}}]$ & 82.94±2.59 & 82.22±2.60 & 85.85±1.66 & 87.12±2.22 \\
        + $[\boldsymbol{x}_{\text{lap}} \mathbin\Vert \boldsymbol{x}_{\text{char}} \mathbin\Vert \boldsymbol{x}_{\text{nbr}}]$ & \textbf{83.49±2.66} & \textbf{82.75±2.75} & \textbf{87.01±2.18} & \textbf{87.19±2.48} \\
        \bottomrule
    \end{tabular}
    }
    \vspace{-1em}
\end{table}

\section{Conclusion}
We present TFM4GAD, a framework adapting Tabular Foundation Models for Graph Anomaly Detection.
By strategically flattening graph-structured data into an augmented tabular format, incorporating Laplacian embeddings, explicit structural metrics, and anomaly-sensitive neighborhood aggregations, we effectively bridge the gap between graph topology and tabular inference.
This approach bypasses expensive training, enabling superior in-context detection performance compared to specialized baselines.
TFM4GAD establishes a new, efficient paradigm for foundation graph anomaly detectors.

\begin{acks}
This work is partially supported by the National Natural Science Foundation of China (62306137) and the Ant Group Research Intern Program.
\end{acks}

\bibliographystyle{ACM-Reference-Format}
\bibliography{main}


\end{document}